\documentclass[conference]{IEEEtran}
\usepackage[british]{babel}
\usepackage{balance}
\usepackage{times}
\usepackage{epsfig}
\usepackage{graphicx}
\usepackage{amsmath}
\usepackage{amssymb}
\usepackage{verbatim}
\usepackage{graphicx,times}
\usepackage{graphics}
\usepackage{xspace}
\usepackage{subfig}
\usepackage{sidecap}
\usepackage{lipsum}          % to automatically generate text
\usepackage{listings}        % to import source code
\usepackage{longtable,lscape}
\usepackage{color}           % For creating coloured text and background
\usepackage{url}
\usepackage{booktabs}
\usepackage{listings}
\usepackage{fancyvrb}
\usepackage{framed}
\usepackage{mdframed}
\usepackage{algpseudocode}
\usepackage{algorithm} %ctan.org\pkg\algorithms
\usepackage{comment}

\ifCLASSINFOpdf
\else
\fi
\definecolor{light-gray}{gray}{0.95}
\begin{document}

\title{Convolutional Neural Networks for Page Segmentation of Historical Document Images}

\author{
\IEEEauthorblockN
{Kai Chen and Mathias Seuret}
\IEEEauthorblockA{
DIVA (Document, Image and Voice Analysis) research group\\
Department of Informatics, University of Fribourg, Switzerland\\
%Bd. de P\'erolles 90, 1700 Fribourg, Switzerland\\ 
Email: \{firstname.lastname\}@unifr.ch}
}

\begin{comment}
\author{
\IEEEauthorblockN
{Kai Chen\IEEEauthorrefmark{1},
Mathias Seuret\IEEEauthorrefmark{1},
Jean Hennebert\IEEEauthorrefmark{1}\IEEEauthorrefmark{2},
and Rolf Ingold\IEEEauthorrefmark{1}}
\IEEEauthorblockA{
\IEEEauthorrefmark{1}DIVA, University of Fribourg, Switzerland, Email: \{firstname.lastname\}@unifr.ch \\
}
%Bd. de P\'erolles 90, 1700 Fribourg, Switzerland\\ 
%Email: \{firstname.lastname\}@unifr.ch}
\IEEEauthorblockA{
\IEEEauthorrefmark{2}University of Applied Sciences, HES-SO//FR, Fribourg, Switzerland, Email: {jean.hennebert@hefr.ch}\\
}
}
\end{comment}

\maketitle

\begin{abstract}
This paper presents a Convolutional Neural Network (CNN) based page segmentation method for handwritten historical document images.
We consider page segmentation as a pixel labeling problem, i.e., each pixel is classified as one of the predefined classes.
Traditional methods in this area rely on carefully hand-crafted features or large amounts of prior knowledge.
In contrast, we propose to learn features from raw image pixels using a CNN.
While many researchers focus on developing deep CNN architectures to solve different problems,
we train a simple CNN with only one convolution layer.
We show that the simple architecture achieves competitive results against other deep architectures on different public datasets. 
Experiments also demonstrate the effectiveness and superiority of the proposed method compared to previous methods.
\end{abstract}


% \fi
\IEEEpeerreviewmaketitle

\section{Introduction}
Page segmentation is an important prerequisite step of document image analysis and understanding. 
The goal is to split a document image into regions of interest. 
Compared to segmentation of machine printed document images, 
page segmentation of historical document images is more challenging due to many variations such as layout structure,
decoration, writing style, and degradation. 
Our goal is to develop a generic segmentation method for handwritten historical document images. 
In this method, we consider the segmentation problem as a pixel-labeling problem, i.e., for
a given document image, each pixel is labeled as one of the predefined classes.

% chen2014robust, 
Some page segmentation methods have been developed recently. 
These methods rely on hand-crafted features~\cite{grana2011automatic, bukhari2012layout, chen2014robust, chen2014page} 
or prior knowledge~\cite{bulacu2007layout, van2011development, panichkriangkrai2013character, gatos2014segmentation}, 
or models that combine hand-crafted features with domain knowledge~\cite{cohen2013robust, asi2014coarse}.
In contrast, in this paper, our goal is to develop a more general method which automatically learns
features from the pixels of document images. 
Elements such as strokes of words, words in sentences, sentences in paragraphs
have a hierarchical structure from low to high levels. 
As these patterns are repeated in different parts of the documents.
Based on these properties, feature learning algorithms can be applied to learn layout information of the document images.

Convolutional Neural Network (CNN) is a kind of feed-forward artificial neural network which shares weights among neurons in the same layer.
By enforcing local connectivity pattern between neurons of adjacent layers, CNN can discover spatially local correlation~\cite{lecun1998gradient}.
With multiple convolutional layers and pooling layers, CNN has achieved many successes in various fields, e.g., 
handwriting recognition~\cite{lecun1989backpropagation},
image classification~\cite{krizhevsky2012imagenet}, 
text recognition in natural images~\cite{wang2012end}, 
and sentence classification~\cite{kim2014convolutional}.

% document image classification~\cite{kang2014convolutional},
% TODO: in our previous
In our previous work~\cite{chen2015page}, an autoencoder to learn features automatically on the training images.
An autoencoder is a feed forward neural network.
The main idea is that by training an autoencoder to reconstruct its input, features can be discovered on the hidden layers.
Then an off-the-shelf classifier can be trained with the learned features to predict pixels into different predefined classes.
By using superpixels as the units of labeling~\cite{chen2016page}, the speed of the method is increased.
In~\cite{chen2016pageb}, a Conditional Random Field (CRF)~\cite{lafferty2001conditional} is applied in order to model the local and contextual information 
jointly to refine the segmentation results which have been achieved in~\cite{chen2016page}.
Following the same idea of~\cite{chen2016pageb}, we consider the segmentation problem as an image patch labeling problem.
The image patches are generated by using superpixels algorithm.
In contrast to~\cite{chen2015page, chen2016page, chen2016pageb}, in this work, we focus on developing an end-to-end method.
We combine feature learning and classifier training into one step.
Image patches are used as input to train a CNN for the labeling task.
During training, the features used to predict labels of the image patches are learned on the convolution layers of the CNN.

While many researchers focus on developing very deep CNN to solving 
various problems~\cite{krizhevsky2012imagenet, zeiler2014visualizing, simonyan2014very, szegedy2015going, he2016deep},
in the proposed method, we train a simple CNN of one convolution layer. 
Experiments on public historical document image datasets show that
despite the simple structure and little tuning of hyperparameters, 
the proposed method achieves comparable results compared to other CNN architectures.

The rest of the paper is organized as follows. Section~\ref{sec:related-work} gives an overview of some related work.
Section~\ref{sec:methodology} presents the proposed CNN for the segmentation task.
Section~\ref{sec:experiments} reports the experimental results and Section~\ref{sec:conclusion} presents the conclusion.

%Gatos et al.~\cite{gatos2014segmentation} propose a text zone and text line segmentation method of handwritten historical documents.
%To detect the main text regions, they first apply a binarization method which is suitable for historical documents as presented in~\cite{gatos2006adaptive}.
%Based on the prior knowledge of the structure of the documents,
%vertical text zones are detected by analyzing vertical rule lines as well as vertical white runs of the document image.
%Text zones are refined by a set of rules which are defined on the connected component level.
%On the detected text zones, a Hough transform based text line segmentation method is used to segment text lines.
%Finally, a set of rules is predefined based on the prior knowledge to refine the text lines.

%Bulacu et al.~\cite{bulacu2007layout} present a method to preserve the ascenders and descenders of text lines of the archive of the cabinet of the Dutch Queen.
%Text lines contain in the main column of the table are extracted by detecting peaks and valleys in the smoothed horizontal projection of the binarized image.
%Then a contour tracing based method is applied to separate text lines.  

\section{Related Work}
\label{sec:related-work}
This section reviews some representative state-of-the-art methods for historical document image segmentation.
Unlike segmentation of contemporary machine printed documents, 
segmentation of handwritten historical documents is more challenging due to the various writing styles,
the rich decoration, degradation, noise, and unrestricted layout.
Therefore, traditional page segmentation method can not be applied to handwritten historical documents directly.
Many methods have been proposed for segmentation of handwritten historical documents,
which largely can be divided into rule based and machine learning based.
%The state-of-the-art methods mostly rely on predefined rules and off-the-shelf classifiers trained with hand-crafted features.

Some methods rely on threshold values predefined based on prior knowledge of the document structure.
Van Phan et al.~\cite{van2011development} use the area of Voronoi
diagram to represent the neighborhood and boundary of connected components (CCs).
By applying predefined rules, characters were extracted by grouping adjacent Voronoi regions.
Panichkriangkrai et al.~\cite{panichkriangkrai2013character} propose a text line and character extraction system of Japanese historical woodblock printed books. 
Text lines are separated by using vertical projection on binarized images. 
To extract kanji characters, rule-based integration is applied to merge or split the CCs.
Gatos et al.~\cite{gatos2014segmentation} propose a text zone and text line segmentation method for handwritten historical documents.
Based on the prior knowledge of the structure of the documents,
vertical text zones are detected by analyzing vertical rule lines and vertical white runs of the document image.
On the detected text zones, a Hough transform based text line segmentation method is used to segment text lines.
All these methods have achieved good segmentation results on specific document datasets.
However, the common limitation is that 
a set of rules have to be carefully defined and document structure is assumed to be observed.
Benefit from the prior knowledge of the structure, the thresholds values are tuned in order to archive good performance.
In other words, due to the generality, the rule-based methods can not be applied to other kinds of historical document images directly. 

%In order to extract pictures on the document images, the CCs are extracted from the binarized images.
%Features such as RGB histogram, enhanced HSV histogram, Gradient Spatial Dependency Matrix are extracted on the components by using a sliding window.
%Then an SVM is trained on the features to classify the components into picture and non-picture.

In order to increase the generality and robustness of page segmentation methods, 
machine learning techniques are employed.
In this case, usually the segmentation problem is considered as a pixel labeling problem.
Feature representation is the key of the machine learning based methods.  
Carefully hand-crafted features are designed in order to train an off-the-shelf classifier on the labeled training set.
Bukhari et al.~\cite{bukhari2012layout} propose a text segmentation method of Arabic historical document images.
They consider the normalized height, foreground area, relative distance, orientation, 
and neighborhood information of the CCs as features.
Then the features are used to train a multilayer perceptron (MLP). 
Finally, the trained MLP is used to classify CCs to relevant classes of text.
Cohen et al.~\cite{cohen2013robust} apply Laplacian of Gaussian on the multi-scale binarized image to extract CCs. 
Based on prior knowledge, appropriate threshold values are chosen in order to remove noise CCs. 
With an energy minimization method and the features, such as bounding box size, area, stroke width, and estimated text lines distance,
each CC is labeled into text or non-text.
Asi et al.~\cite{asi2014coarse} propose a two-steps segmentation method of Arabic historical document images. 
They first extract the main text area with Gabor filters.
Then the segmentation is refined by minimization an energy function.
Compared to the rule based methods, 
the advantage of the machine learning based methods is less prior knowledge is needed.
However, the existing machine learning based methods rely on hand-crafted feature engineering 
and to obtain appropriate hand-crafted features for specific tasks is cumbersome.

%Grana et al.~\cite{grana2011automatic} propose a texture analysis based method to segment textual and pictorial regions on historical document images.
%First, a central symmetric matrix is obtained by applying the texture features which are extracted with an autocorrelation function of the grayscale image. 
%Then the image is divided into square blocks.
%With a Support Vector Machine (SVM) trained on the texture features, blocks are classified into text and non-text blocks.

\section{Methodology}
\label{sec:methodology}
In order to create general page segmentation method without using any prior knowledge of the layout structure of the documents,
we consider the page segmentation problem as a pixel labeling problem.
We propose to use a CNN for the pixel labeling task. 
The main idea is to learn a set of feature detectors and train a nonlinear classifier on the features extracted by the feature detectors.
With the set of feature detectors and the classifier, pixels on the unseen document images can be classified into different classes.

\subsection{Preprocessing}
\label{sec:preprocessing}
In order to speed up the pixel labeling process, for a given document image, 
we first applying a superpixel algorithm to generate superpixels.
A superpixel is an image patch which contains the pixels belong to the same object.
Then instead of labeling all the pixels, we only label the center pixel of each superpixel 
and the rest pixels in that superpixel are assigned to the same label.    
The superiority of the superpixel labeling approach over the pixel labeling approach for the page segmentation task has been demonstrated in~\cite{chen2016page}.
Based on the previous work~\cite{chen2016page}, the simple linear iterative clustering (SLIC) algorithm is applied 
as a preprocessing step to generate superpixels for given document images.

\subsection{CNN Architecture}
\label{sec:cnn-architecture}
\begin{figure}[!tb]
	\vspace{-10pt}
	\centering
	\includegraphics[width=0.9\linewidth]{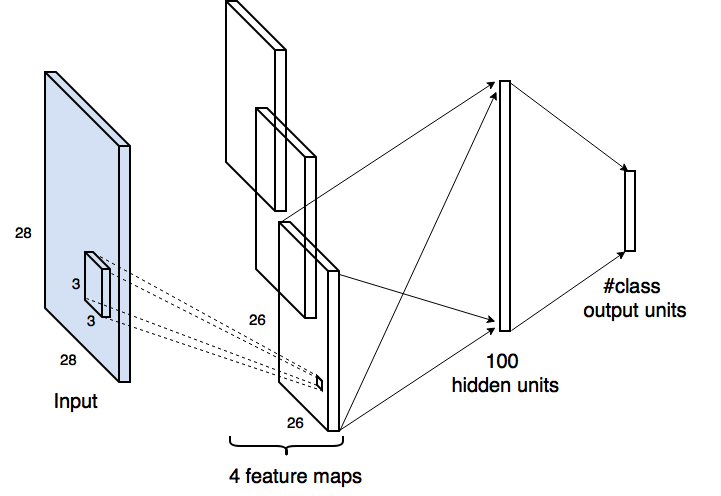}
	\caption{The architecture of the proposed CNN}
  	\label{fig:cnn}
	\vspace{-20pt}
\end{figure}

The architecture of our CNN is given in Figure~\ref{fig:cnn}. 
The structure can be summarized as $28 \times 28 \times 1 - 26 \times 26 \times 4 - 100 - M$, where $M$ is the number of classes.
The input is a grayscale image patch. The size of the image patch is $28 \times 28$ pixels. 
Our CNN architecture contains only one convolution layer which consists of $4$ kernels. 
The size of each kernel is $3 \times 3$ pixels.
Unlike other traditional CNN architecture, the pooling layer is not used in our architecture.  
Then one fully connected layer of $100$ neurons follows the convolution layer.
The last layer consists of a logistic regression with softmax which outputs the probability of each class, such that

\begin{equation}\label{eq:logistic-regression}
	P(y=i | x, W_1, \cdots, W_M, b_1, \cdots, b_M ) = \frac{e^{W_{i}x + b_i}}{\sum_{j=1}^{M} e^{W_{j}x + b_j}},
\end{equation}
where $x$ is the output of the fully connected layer, $W_i$ and $b_i$ are the weights and biases of the $i^{th}$ neuron in this layer, and $M$ is the number of the classes.
The predicted class $\hat{y}$ is the class which has the max probability, such that

\begin{equation}\label{eq:prediction}
	\hat{y} = \operatorname*{arg\,max}_i P(y=i | x, W_1, \cdots, W_M, b_1, \cdots, b_M).
\end{equation}

In the convolution and fully connected layers of the CNN, Rectified Linear Units (ReLUs)~\cite{nair2010rectified} are used as neurons.
An ReLU is given as:
\begin{equation}\label{eq:relu}
	f(x) = \max(0, x),
\end{equation}
where $x$ is the input of the neuron.
The superiority of using ReLUs as neurons in CNN over traditional sigmoid neurons is demonstrated in~\cite{krizhevsky2012imagenet}. 

% Feature extraction
%++++++++++++++++++++++++++++++++++++++++++++++++++++++++++++++++++++++++++++
%++++++++++++++++++++++++++++++++++++++++++++++++++++++++++++++++++++++++++++

% we take image patches centred on the superpixels of the training images as input.
\subsection{Training}
\label{sec:training}
To train the CNN, for each superpixel, we generate a patch which is centred on that superpixel.
The patch is considered as the input of the network.
The size of each patch is $28 \times 28$ pixels.
The label of each patch is its center pixel's label.
The patches of the training images are used to train the network.

In the CNN, the cost function is defined as the cross-entropy loss, such that

\begin{equation}\label{eq:cross-entropy}
	\mathcal{L}(X, Y) = -\frac{1}{n} \sum_{i=1}^{n} (\ln a(x^{(i)}) + (1 - y^{(i)}) \ln (1 - a(x^{(i)}))),
\end{equation}
where $X = \{ x^{(1)}, \cdots, x^{(n)} \}$ is the training image patches 
and $Y = \{ y^{(1)}, \cdots, y^{(n)} \}$ is the corresponding set of labels. 
The number of training image patches is $n$. For each $x^{(i)}$, $a(x^{(i)})$ is the output of the CNN as defined in Eq.~\ref{eq:logistic-regression}.
The CNN is trained with Stochastic Gradient Descent (SGD) with the dropout~\cite{srivastava2014dropout} technique.
The goal of dropout is to avoid overfitting by introducing random noise to training samples.
Such that during the training, the outputs of the neurons are masked out with the probability of $0.5$.

\section{Experiment}
\label{sec:experiments}
Experiments on six public handwritten historical document image datasets are conducted.

%The \emph{Parzival} dataset consits of the pages written by three writers.
%These pages are taken from a medieval German manuscript from the $13$th century that contains the epic poem Parzival by Wolfram von Eschenbach.
%The \emph{St. Gall} dataset consists the manuscripts from a medieval manuscript written in Latin.

% while the format suffers from many degradations.
\subsection{Datasets}
\label{sec:dataset}

%Six historical document datasets are used for the experiments.
The datasets are of very different nature.
The \emph{G. Washington} dataset consists of the pages written in English with ink on paper and the images are in gray levels.
The other two datasets, i.e., \emph{Parzival} and \emph{St. Gall} datasets consist of images of manuscripts written with ink on parchment 
and the images are in color.
The \emph{Parzival} dataset consits of the pages written by three writers in the $13$th century.
The \emph{St. Gall} dataset consists the manuscripts from a medieval manuscript written in Latin.
The details of the ground truth are presented in~\cite{chen2015ground}.

% using Carolingian minuscule script
% The manuscripts have quite similar layout features.
Three new datasets with more complex layout have been recently created~\cite{simistira16}.
The \emph{CB55} dataset consists of manuscripts from the $14$th century which are written in Italian and Latin languages by one writer.
The \emph{CSG18} and \emph{CSG863} datasets consist of manuscripts from the $11$th century which are written in Latin language.
The number of the writers of the two datasets is not specified.
The details of the three datasets are presented in~\cite{simistira16}.

In the experiments, all images are scaled down with a scaling factor $2^{-3}$. 
Table~\ref{tab:dataset} gives the details of training, test, and validation sets of the six datasets.

% dataset
%++++++++++++++++++++++++++++++++++++++++++++++++++++++++++++++++++++++++++++
\begin{table}[!tb]
\caption{Details of training, test, and validation sets. $TR$, $TE$, and $VA$ denotes the training, test, and validation sets respectively.}
\vspace{-5pt}
	\centering
%\scalebox{0.88} {
\begin{tabular}
	{l c c c c}
    %{ p{2cm}	                      p{2cm}		                p{0.5cm}		             p{0.5cm}	 p{0.5cm}}
	%\hline
	\toprule
			 			   		& image size (pixels) & $|TR|$	 & $|TE|$ & $|VA|$ \\  
    \hline
	\emph{G. Washington}  	& $2200 \times 3400$  	& 10 	 & 5	      & 4 \\
	\emph{St. Gall}  		& $1664 \times 2496$  	& 20 	 & 30	  & 10 \\
	\emph{Parzival}  	    & $2000 \times 3008$  	& 20 	 & 13	  & 2  \\
	\hline
	\emph{CB55}  			& $4872 \times 6496$  	& 20 	 & 10	  & 10 \\
	\emph{CSG18}  			& $3328 \times 4992$  	& 20 	 & 10	  & 10 \\
	\emph{CSG863}  	    		& $3328 \times 4992$  	& 20 	 & 10	  & 10  \\
	\bottomrule
\end{tabular}
%}
\label{tab:dataset}
\vspace{-15pt}
\end{table}
%++++++++++++++++++++++++++++++++++++++++++++++++++++++++++++++++++++++++++++

\subsection{Metrics}
\label{sec:metrics}
To evaluate methods of page segmentation for historical document images, 
the most used metrics are precision, recall, and pixel level accuracy.
In contrast, besides of the standard metrics, 
we adapt the metrics which are well defined and has been widely used from common semantic segmentation and scene parsing evaluations.
The metrics are variations on pixel accuracy and region intersection over union (IU).
They have been proposed in~\cite{long2015fully}.  
Consequently, the metrics used in the experiments are: pixel accuracy, mean pixel accuracy, mean IU, and frequency weighted IU (f.w. IU).

In order to obtained the metrics, we define the variables:
\begin{itemize}
	\item $n_{c}$: the number of classes.
	
	\item $n_{ij}$: the number of pixels of class $i$ predicted to belong to class $j$. For class $i$:
		\begin{itemize}
			\item $n_{ii}$: the number of correctly classified pixels (true positives).
			\item $n_{ij}$: the number of wrongly classified pixels (false positives).
			\item $n_{ji}$: the number of wrongly not classified pixels (false negatives).
		\end{itemize}			
	
	\item $t_{i}$: the total number of pixels in class $i$, such that 
	\begin{equation}
		t_i = \sum_j n_{ji}.
	\end{equation}	
	
\end{itemize}

With the defined variables, we can compute:
\begin{itemize}
	\item pixel accuracy: 
	\begin{equation}
		acc = \frac{\sum_i n_{ii}}{\sum_i t_i} .
	\end{equation}
	
	% ~\footnote{Note that the mean accuracy is the mean recall.}
	\item mean accuracy: 
	\begin{equation}
		acc_{mean} = \frac{1}{n_{c}} \times \sum_i \frac{n_{ii}}{t_i} .
	\end{equation}
	
	\item mean IU: 
	\begin{equation}
		iu_{mean} = \frac{1}{n_{c}} \times \sum_i \frac{n_{ii}}{t_i + \sum_j n_{ji} - n_{ii}} .
	\end{equation}
	
	\item f.w. IU: 
	\begin{equation}
		iu_{weighted} = \frac{1}{\sum_k t_k} \times \sum_i \frac{t_i \times n_{ii}}{t_i + \sum_j n_{ji} - n_{ii}} .
	\end{equation}
\end{itemize}

% TODO: our previous methods
\subsection{Evaluation}
\label{sec:evaluation}
We compare the proposed method to our previous methods~\cite{chen2016page, chen2016pageb}.
Similar to the proposed method, superpixels are considered as the basic unit of labeling.
In~\cite{chen2016page}, the features are learned on randomly selected grayscale image patches with a stacked convolutional autoencoder in an unsupervised manner.
Then the features and the labels of the superpixels are used to train a classifier. 
With the trained classifier, superpixels are classified into different classes.
In~\cite{chen2016pageb}, a Conditional Random Field (CRF) is applied in order to model the local and contextual information jointly for the superpixel labeling task.
The trained classifier in~\cite{chen2016page} is considered as the local classifier in~\cite{chen2016pageb}.
Then the local classifier is used to train a contextual classifier which takes the output of the local classifier as input and output the scores of given labels.
With the local and contextual classifiers, a CRF is trained to label the superpixels of a given image.
In the experiments, we use a multilayer perceptron (MLP) as the local classifier in~\cite{chen2016page, chen2016pageb} 
and another MLP as the contextual classifier in~\cite{chen2016pageb}.
Simple Linear Iterative Clustering algorithm (SLIC)~\cite{achanta2012slic} is applied to generate the superpixels.
The superiority of SLIC over other superpixel algorithms is demonstrated in~\cite{chen2016page}. 
In the experiments, for each image, $3000$ superpixels are generated.

\begin{figure*}[!t]
    %\vspace{-10pt}
    \centering
     %\label{fig:307}
     %\frame{\includegraphics[width=0.15\textwidth]{figures/ground-truth/307.png}} \, 
     %\label{fig:307-ground-truth}
     %\frame{\includegraphics[width=0.15\textwidth]{figures/ground-truth/307_200.png}} \,     
     %\label{fig:307-pixel}
     %\frame{\includegraphics[width=0.15\textwidth]{figures/segmentation/local-mlp/307_250.png}} \,         
     %\label{fig:307-local-mlp}
     %\frame{\includegraphics[width=0.15\textwidth]{figures/segmentation/crf/307_250.png}} \,
     %\label{fig:307-crf}
     %\frame{\includegraphics[width=0.15\textwidth]{figures/segmentation/cnn/307_250.png}} \\
     %\vspace{2pt}
     \label{fig:d-144-input}
     \frame{\includegraphics[width=0.16\textwidth]{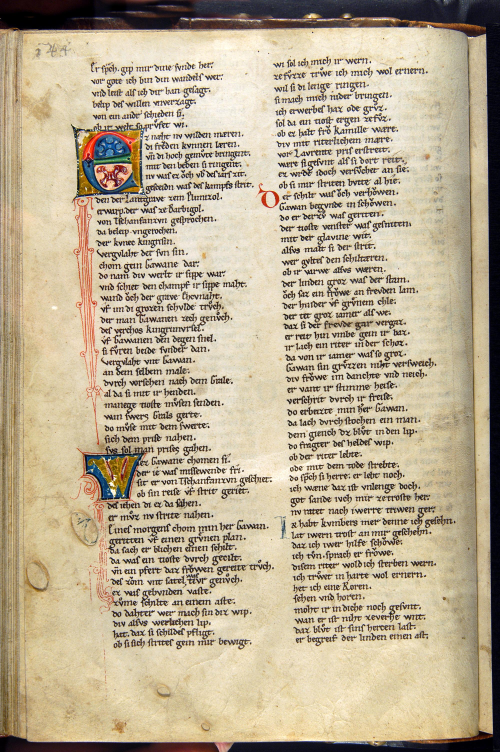}} \, 
     \label{fig:d-144-ground-truth}
     \frame{\includegraphics[width=0.16\textwidth]{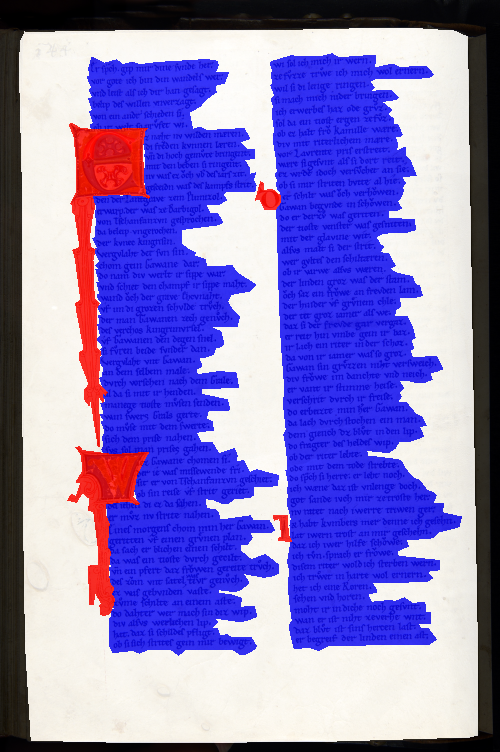}} \,     
     \label{fig:d-144-pixel}
     \frame{\includegraphics[width=0.16\textwidth]{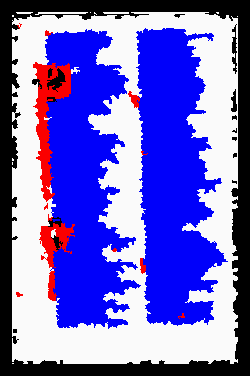}} \,         
     \label{fig:d-144-local-mlp}
     \frame{\includegraphics[width=0.16\textwidth]{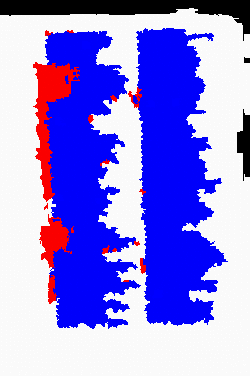}} \,
     \label{fig:d-144-crf}
     \frame{\includegraphics[width=0.16\textwidth]{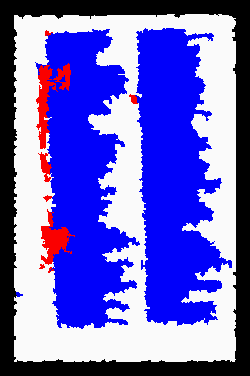}} \\
     \vspace{2pt}
     %\label{fig:csg562-029-input}
     %\frame{\includegraphics[width=0.15\textwidth]{figures/ground-truth/csg562-029.png}} \, 
     %\label{fig:csg562-029-ground-truth}
     %\frame{\includegraphics[width=0.15\textwidth]{figures/ground-truth/csg562-029_200.png}} \,     
     %\label{fig:csg562-029-pixel}
     %\frame{\includegraphics[width=0.15\textwidth]{figures/segmentation/local-mlp/csg562-029_250.png}} \,         
     %\label{fig:csg562-029-local-mlp}
     %\frame{\includegraphics[width=0.15\textwidth]{figures/segmentation/crf/csg562-029_250.png}} \,
     %\label{fig:csg562-029-crf}
     %\frame{\includegraphics[width=0.15\textwidth]{figures/segmentation/cnn/csg562-029_250.png}} \\
     %\vspace{2pt}
     \label{fig:cb55}
     \frame{\includegraphics[width=0.16\textwidth]{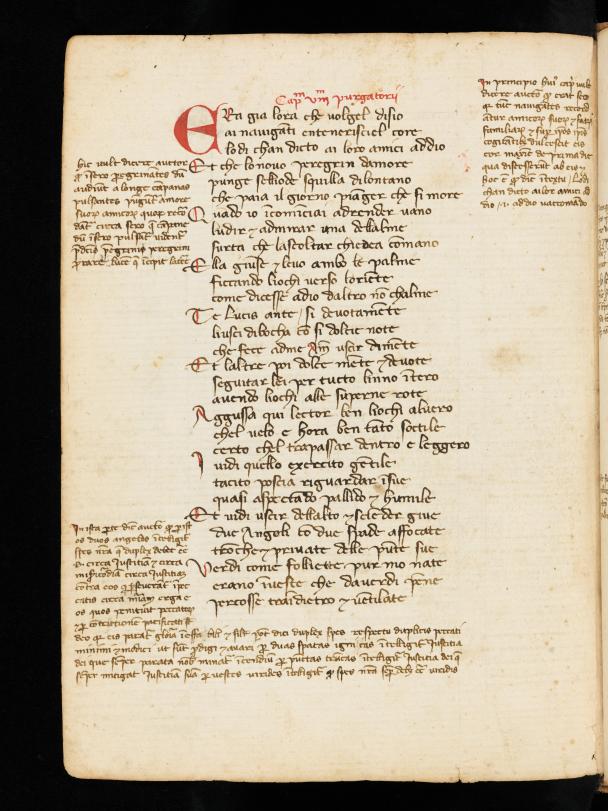}} \, 
     \label{fig:ch-crf-cb55-ground-truth}
     \frame{\includegraphics[width=0.16\textwidth]{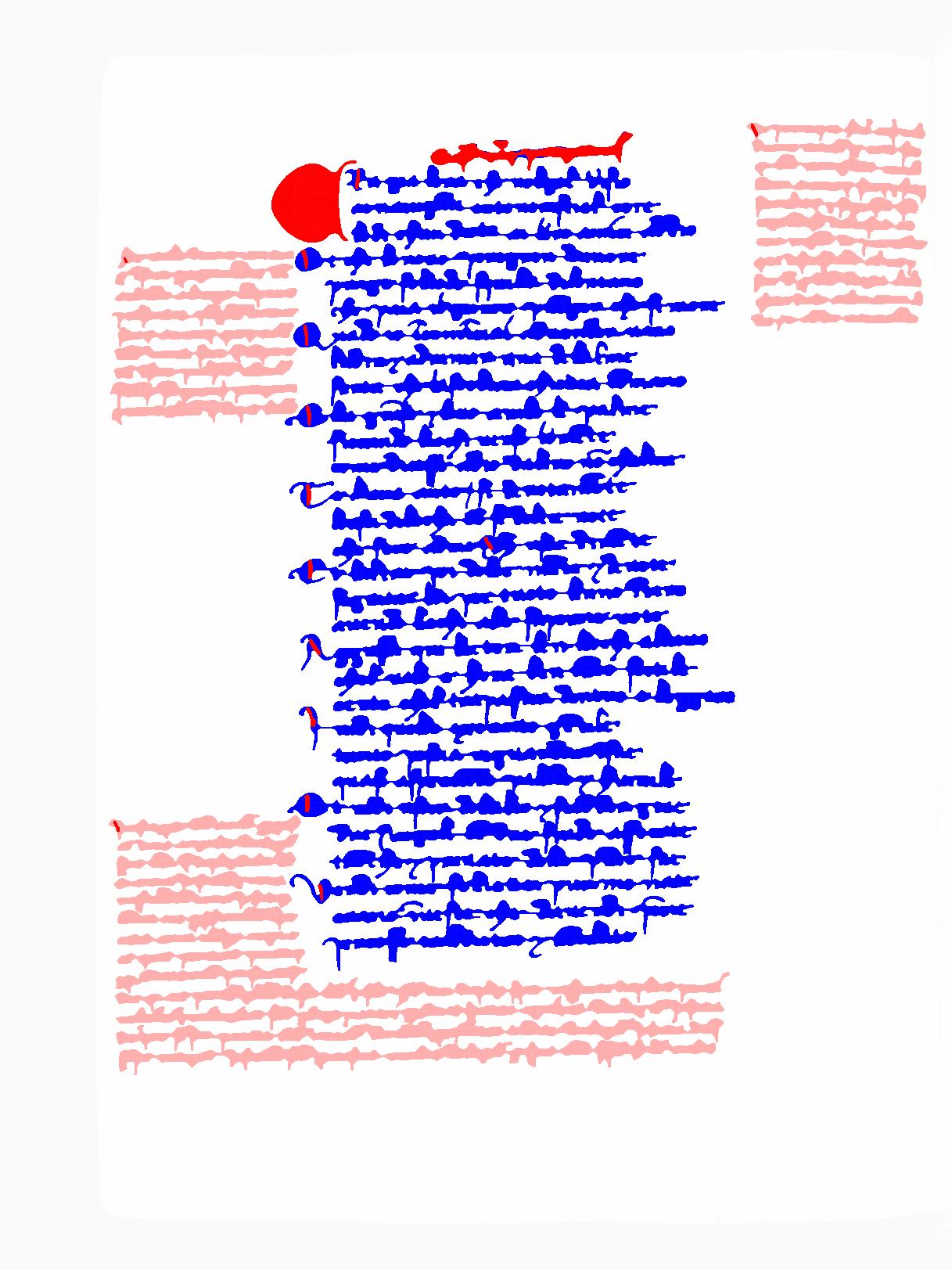}} \,     
     \label{fig:cb55-pixel}
     \frame{\includegraphics[width=0.16\textwidth]{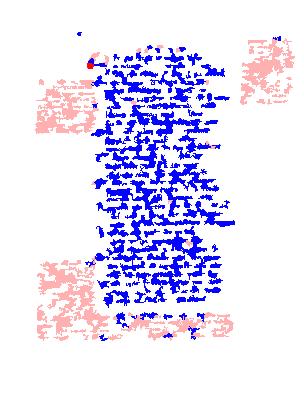}} \,         
     \label{fig:cb55-local-mlp}
     \frame{\includegraphics[width=0.16\textwidth]{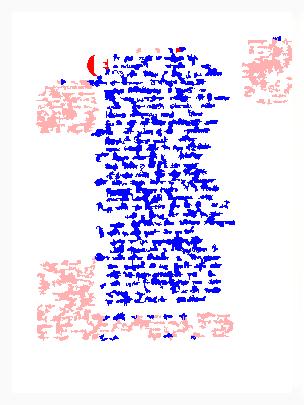}} \,
     \label{fig:cb55-crf}
     \frame{\includegraphics[width=0.16\textwidth]{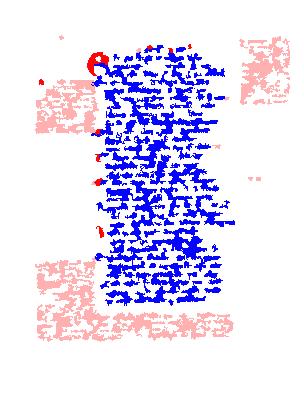}} \\
     \vspace{2pt}
     %\label{fig:csg18-input}
     %\frame{\includegraphics[width=0.15\textwidth]{figures/ground-truth/e-codices_csg-0018_098_max.jpg}} \, 
     %\label{fig:csg18-ground-truth}
     %\frame{\includegraphics[width=0.15\textwidth]{figures/ground-truth/e-codices_csg-0018_098_max_gt.jpg}} \,     
     %\label{fig:csg18-pixel}
     %\frame{\includegraphics[width=0.15\textwidth]{figures/segmentation/local-mlp/e-codices_csg-0018_098_max.jpg}} \,         
     %\label{fig:csg18-local-mlp}
     %\frame{\includegraphics[width=0.15\textwidth]{figures/segmentation/crf/e-codices_csg-0018_098_max.jpg}} \,
     %\label{fig:csg18-crf}
     %\frame{\includegraphics[width=0.15\textwidth]{figures/segmentation/cnn/e-codices_csg-0018_098_max_250.jpg}} \\
     %\vspace{2pt}
     \label{fig:csg863-input}
     \frame{\includegraphics[width=0.16\textwidth]{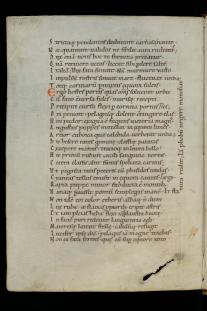}} \, 
     \label{fig:csg863-ground-truth}
     \frame{\includegraphics[width=0.16\textwidth]{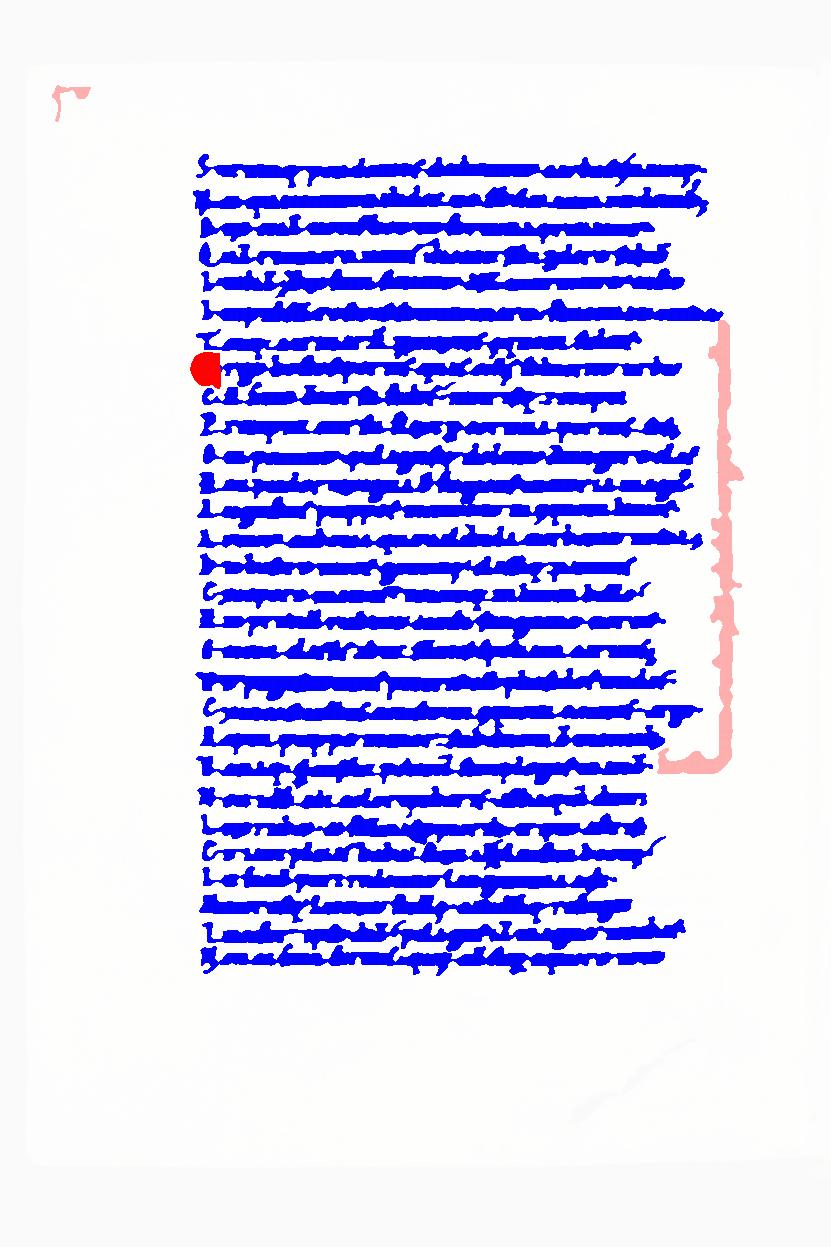}} \,          
     \label{fig:csg863-pixel}
     \frame{\includegraphics[width=0.16\textwidth]{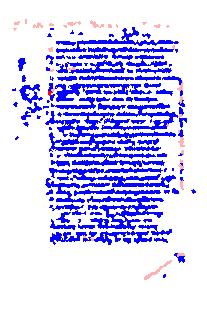}} \,         
     \label{fig:csg863-local-mlp}
     \frame{\includegraphics[width=0.16\textwidth]{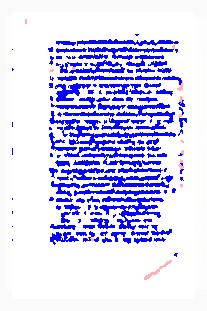}} \, 
     \label{fig:csg863-crf}
     \frame{\includegraphics[width=0.16\textwidth]{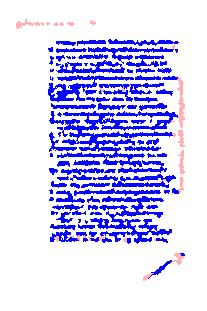}} \\
\caption{
Segmentation results on the \emph{Parzival}, \emph{CB55}, and \emph{CSG863} datasets from top to bottom respectively.
The colors: black, white, blue, red, and pink are used to represent: \emph{periphery}, \emph{page}, \emph{text}, \emph{decoration}, and \emph{comment} respectively.
The columns from left to right are: input, ground truth, segmentation results of the local MLP, CRF, and CNN respectively.}
  \label{fig:performance-comparison}
  %\vspace{10pt}
\end{figure*}

Table~\ref{tab:performance-comparison-1} reports the pixel accuracy, mean pixel accuracy, mean IU, and f.w. IU of the three methods.
It is shown that the proposed CNN outperforms the previous method.
Figure~\ref{fig:performance-comparison} gives the segmentation results of the three methods.
We can see that visually the CNN achieve more accurate segmentation results compared to other methods.

% performance comparison
%++++++++++++++++++++++++++++++++++++++++++++++++++++++++++++++++++++++++++++
\begin{table*}[!t]
    \caption{Performance (in percentage) of superpixel labeling with only local MLP, CRF, and the proposed CNN.}
    \centering
%\hspace*{-1cm}
\begin{tabular}
    { l c c c c c c c c c c c c }
    %{ p{2cm}                          p{2cm}                        p{0.5cm}                     p{0.5cm}     p{0.5cm}}
    %\hline
    \toprule
    %\hline    
    %\multicolumn{13}{l}{{$\alpha=2^{-3}$}} \\
    %\hline
                         & \multicolumn{4}{c}{\emph{G. Washington}}  &  \multicolumn{4}{c}{\emph{Parzival}}        &  \multicolumn{4}{c}{\emph{St.Gall}}  \\  
    \hline
          & {pixel}     & {mean}     & {mean}    & {f.w.}  & {pixel}     & {mean}     & {mean}    & {f.w.} & {pixel}     & {mean}     & {mean}    & {f.w.} \\
          & {acc.}     & {acc.}     & {IU}        & {IU}        & {acc.}     & {acc.}     & {IU}        & {IU}     & {acc.}     & {acc.}     & {IU}        & {IU}\\    
    %\hline    
    %\multicolumn{13}{l}{\textbf{Color}} \\
    %\hline
    {Local MLP~\cite{chen2016page}}    & 87 & 89 & 75 & 83 			& 91 & 64 & 58 & 86 		& 95 & 89 & 84 & 92 \\
    {CRF~\cite{chen2016pageb}}       	  & \textbf{91} & 90 & 76 & 85 	& 93 & 70 & 63 & 88 		& 97 & 88 & 84 & 94 \\
    {CNN}         & \textbf{91} & \textbf{91} & \textbf{77} & \textbf{86} & \textbf{94} & \textbf{75} & \textbf{68} & \textbf{89}  & \textbf{98} & \textbf{90} & \textbf{87} & \textbf{96} \\
    {CNN (max pooling)}  & \textbf{91} & 90 & \textbf{77} & \textbf{86} & \textbf{94} & \textbf{75} & \textbf{68} & \textbf{89}  & \textbf{98} & \textbf{90} & \textbf{87} & \textbf{96} \\
    %\hline
    \toprule
    & \multicolumn{4}{c}{\emph{CB55}}  &  \multicolumn{4}{c}{\emph{CSG18}}        &  \multicolumn{4}{c}{\emph{CSG863}}  \\  
    \hline
          & {pixel}     & {mean}     & {mean}    & {f.w.}  & {pixel}     & {mean}     & {mean}    & {f.w.} & {pixel}     & {mean}     & {mean}    & {f.w.} \\
          & {acc.}     & {acc.}     & {IU}        & {IU}        & {acc.}     & {acc.}     & {IU}        & {IU}     & {acc.}     & {acc.}     & {IU}        & {IU}\\    
    %\hline    
    %\multicolumn{13}{l}{\textbf{Color}} \\
    %\hline
    {Local MLP~\cite{chen2016page}}    & 83 & 53 & 42 & 72 		& 83 & 49 & 39 & 73 			& 84 & 54 & 42 & 74 \\
    {CRF~\cite{chen2016pageb}}     	   & 84 & 53 & 42 & 75 		& 86 & 47 & 37 & 77 			& 86 & 51 & 42 & 78 \\
    {CNN}          & \textbf{86} & 59 & 47 & \textbf{77}	& \textbf{87} & \textbf{53} & 41 & 79  & \textbf{87} & \textbf{58} & \textbf{45} & \textbf{79} \\
    {CNN (max pooling)} & \textbf{86} & \textbf{60} & \textbf{48} & \textbf{77}	& \textbf{87} & \textbf{53} & \textbf{42} & \textbf{80}  & \textbf{87} & 57 & \textbf{45} & \textbf{79} \\
    \bottomrule
\end{tabular}
\label{tab:performance-comparison-1}
   \vspace{-10pt}
\end{table*}
%++++++++++++++++++++++++++++++++++++++++++++++++++++++++++++++++++++++++++++

%\subfigure[]{}
%++++++++++++++++++++++++++++++++++++++++++++++++++++++++++++++++++++++++++++
%++++++++++++++++++++++++++++++++++++++++++++++++++++++++++++++++++++++++++++

%Segmentation results on the \emph{G.Washington}, \emph{Parzival}, \emph{St.Gall}, 
%\emph{CB55}, \emph{CSG18}, and \emph{CSG863} datasets from top to bottom respectively.
%The colors: black, white, blue, red, and pink are used to represent: \emph{periphery}, \emph{page}, \emph{text}, \emph{decoration}, and \emph{comment} respectively.
%The columns from left to right are: input, ground truth, segmentation results of the local MLP, CRF, and CNN respectively

\subsection{Max Pooling}
\label{sec:max-pooling}
Pooling is a widely used technology in CNN. 
Max pooling is the most common type of pooling which is applied in order to 
reduce spatial size of the representation to reduce the number of parameters of the network.
In order to show the impact of max pooling for the segmentation task.
We add a max pooling layer after the convolution layer.
The pooling size is $2 \times 2$ pixels.
Table~\ref{tab:performance-comparison-1} reports the performance of the CNN with a max pooling layer.
We can see that only on the \emph{CB55} dataset, the mean pixel accuracy and mean IU are slightly improved.
In general, adding a max pooling layer does not improve the performance of the segmentation task.
The reason is that for some computer vision problems, e.g., object recognition and text extraction in natural images, 
the exact location of a feature is less important than its rough location relative to other features.
However, for a given document image, 
to label a pixel in the center of a patch, it is not sufficient to know if there is text somewhere in that patch, but also the location of the text.
Therefore, the exact location of a feature is helpful for the page segmentation task.
 
% f.w. IU on different number of convolutional layers
%++++++++++++++++++++++++++++++++++++++++++++++++++++++++++++++++++++++++++++
\begin{figure}[!tb]
	%\vspace{-10pt}
	\centering
	\includegraphics[width=0.85\linewidth]{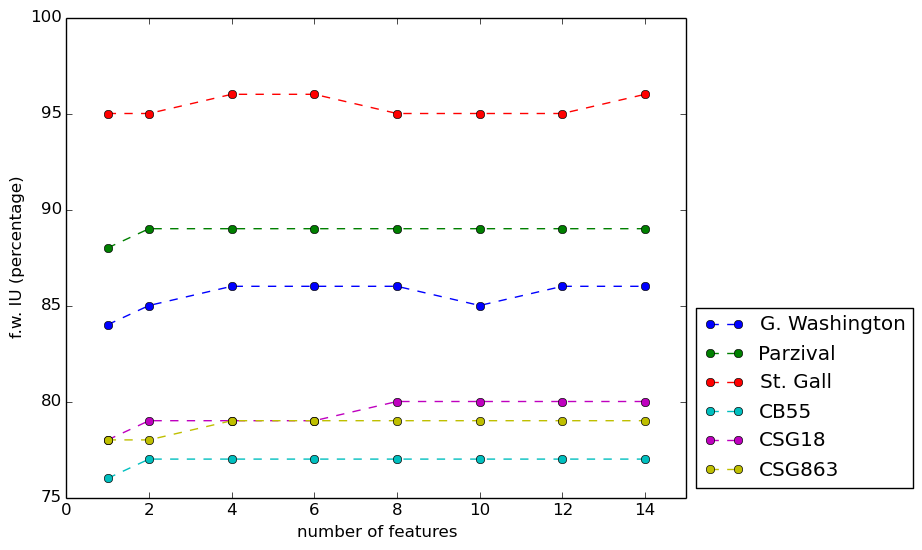}
	\caption{f.w. IU of the one convolution layer CNN on different number of filters.}
  	\label{fig:performance-nb-features}
	\vspace{-10pt}
\end{figure} 																										
%++++++++++++++++++++++++++++++++++++++++++++++++++++++++++++++++++++++++++++

\subsection{Number of Kernels}
\label{sec:nb-filters}
In order to show the impact of the number of kernels of the convolution layer on the segmentation task.
We define the number of kernels as $K$.
In the experiments, we set $K \in \{ 1, 2, 4, 6, 8, 10, 12, 14 \}$.
Figure~\ref{fig:performance-nb-features} reports the f.w. IU of the one convolution layer CNN with different number of kernels.
We can see that except on the \emph{CS18} dataset, when $K \geq 4$ the performance is not improved.

\subsection{Number of Layers}
\label{sec:nb-layers}
\begin{figure}[!tb]
	%\vspace{-10pt}
	\centering
	\includegraphics[width=0.85\linewidth]{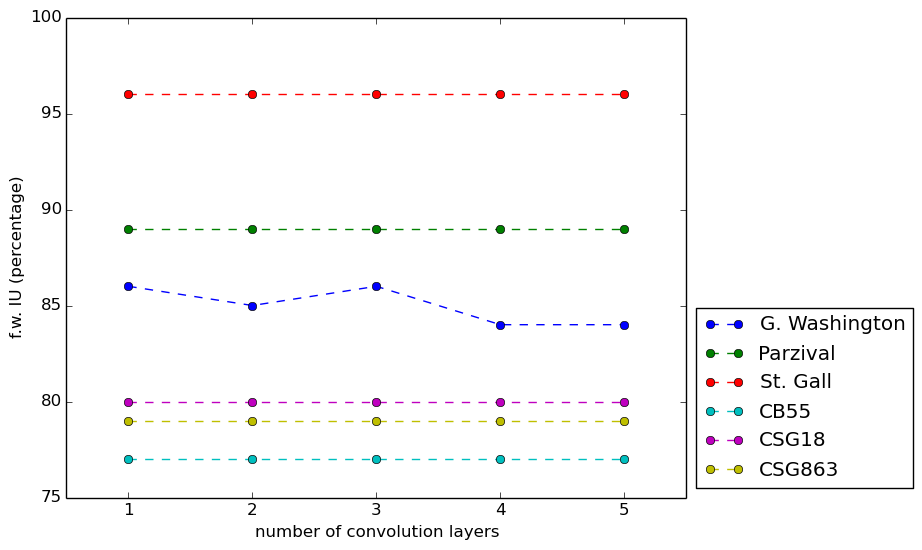}
	\caption{f.w. IU of the CNN on different number of convolutional layers.}
  	\label{fig:performance-nb-conv-layers}
	\vspace{-10pt}
\end{figure}

In order to show the impact of the number of convolutional layers on the page segmentation task.
We incrementally add convolutional layers, such that there is two more kernels on the current layer than the previous layer.
Figure~\ref{fig:performance-nb-conv-layers} reports the f.w. IU of the CNN with different number of convolutional layers.
It is show that the number of layers does not affect the performance of the segmentation task.
However, on the \emph{G. Washington} dataset, with more layers, the performance is degraded slightly.
The reason is that compared to other datasets, the \emph{G. Washington} dataset has fewer training images. 

% f.w. IU on different number of convolutional layers
%++++++++++++++++++++++++++++++++++++++++++++++++++++++++++++++++++++++++++++
%++++++++++++++++++++++++++++++++++++++++++++++++++++++++++++++++++++++++++++

\subsection{Number of Training Images}
\label{sec:training-images}
\begin{figure}[!tb]
	%\vspace{-10pt}
	\centering
	\includegraphics[width=0.85\linewidth]{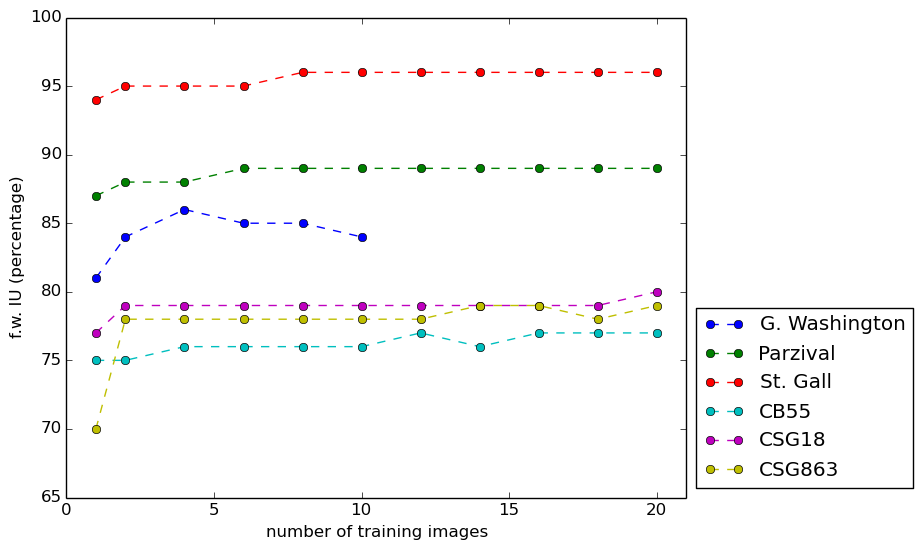}
	\caption{f.w. IU of the CNN on different number of training images.}
  	\label{fig:performance-nb-train-images}
	\vspace{-10pt}
\end{figure}

In order to show the performance under different amount of training images.
For each dataset, we choose $N$ images in the training set to train the CNN.
For each experiment, the number of batches is set to $5000$.
Figure~\ref{fig:performance-nb-train-images} reports the f.w. IU under different values of $N$, such that $N \in \{1, 2, 4, 8, 10, 12, 14, 16, 18, 20 \}$\footnote{In the \emph{G. Washington} dataset, there is $10$ training images. Therefore, $N \in \{1, 2, 4, 8, 10 \}$}.
We can see that in general, when $N > 2$, the performance is not improved.
However, on the \emph{G. Washington} dataset, with more training images, the performance is degraded slightly.
The reason is that compared to the other datasets, on the \emph{G. Washington} dataset the pages are more varied
and the ground truth is less constant.

% f.w. IU on different number of training images
%++++++++++++++++++++++++++++++++++++++++++++++++++++++++++++++++++++++++++++
%++++++++++++++++++++++++++++++++++++++++++++++++++++++++++++++++++++++++++++

\subsection{Run Time}
\label{sec:run-time}																																					
The proposed CNN is implemented with the python library Theano~\cite{bergstra2010theano}.
The experiments are performed on a PC with an Intel Core i7-3770 3.4 GHz processor and 16 GB RAM. 
On average, for each image, the CNN takes about $1$ second processing time.
The superpixel labeling method~\cite{chen2016page} and CRF model~\cite{chen2016pageb}	take about $2$ and $5$ seconds respectively.

\section{Conclusion}
\label{sec:conclusion}
In this paper, we have proposed a convolutional neural network (CNN) for page segmentation of handwritten historical document images.
In contrast to traditional page segmentation methods which rely on off-the-shelf classifiers trained with hand-crafted features,
the proposed method learns features directly from image patches. 
Furthermore, feature learning and classifier training are combined into one step. 
Experiments on public datasets show the superiority of the proposed method over previous methods. 
While many researchers focus on applying very deep CNN architectures for different tasks,
we show that with the simple one convolution layer CNN, we achieve comparable performance compared to other network architectures.


%\section*{Acknowledgment}



\bibliographystyle{IEEEtran}
\bibliography{bare_conf}
%\begin{thebibliography}{1}


%\end{thebibliography}

\end{document}